\documentclass{egpubl}
\usepackage{eg2025}
 
\ShortPresentation      
\usepackage[T1]{fontenc}
\usepackage{dfadobe}  

\usepackage{cite}  
\BibtexOrBiblatex
\electronicVersion
\PrintedOrElectronic
\ifpdf \usepackage[pdftex]{graphicx} \pdfcompresslevel=9
\else \usepackage[dvips]{graphicx} \fi

\usepackage{egweblnk} 
\usepackage{cleveref}

\title[PhyDeformer]%
      {PhyDeformer: High-Quality Non-Rigid Garment Registration with Physics-Awareness}

\author[Boyang Yu \& Frederic Cordier \& Hyewon Seo]
{\parbox{\textwidth}{\centering Boyang Yu $^{1}$\orcid{0000-0002-0934-610X} \, Frederic Cordier $^{2}$\orcid{0000-0003-0675-5431}
        and Hyewon Seo$^{1}$\orcid{0000-0001-8851-0256} 
        }    
        \\
{\parbox{\textwidth}{\centering $^1$ ICube laboratory, CNRS–University of
Strasbourg, France \\
         $^2$ IRIMAS, University of Haute-Alsace, France%
       }
}
}

\usepackage{soul}

\begin{document}


\maketitle
\begin{abstract}
    We present PhyDeformer, a new deformation method for high-quality garment mesh registration. It operates in two phases: In the first phase, a garment grading is performed to achieve a coarse 3D alignment between the mesh template and the target mesh, accounting for proportional scaling and fit (e.g. length, size). Then, the graded mesh is refined to align with the fine-grained details of the 3D target through an optimization coupled with the Jacobian-based deformation framework. Both quantitative and qualitative evaluations on synthetic and real garments highlight the effectiveness of our method.


\end{abstract}  
\section{Introduction}

Real 3D garment data is becoming increasingly prevalent in virtual clothing, enhancing realism in many applications such as gaming, film, fashion, and virtual try-on. The integration of such real-world garment into virtual clothing is often challenging due to the large variations and high-frequency details present in realistic garments geometries. In this paper, we propose a deformation method tailored for high-quality garment mesh registration. Our two-stage mesh deformation approach, named PhyDeformer, effectively performs non-rigid registration to capture a wide variety of garment shapes, handling both large modifications (e.g., proportional resizing) and intricate details, such as folds and wrinkles. In the first stage, garment grading is performed to achieve a coarse 3D alignment between the template and the target mesh, accounting for proportional scaling and fit. In the second stage, the graded garment mesh undergoes refinement using Jacobian-based deformation, guided by both the reconstruction loss and the physics-based constraints. As a result, we obtain a resulting mesh that fits the target well while maintaining physical plausibility.

\section{Related work}

Most existing methods compute vertex-wise displacement map to align a single-layer mesh \cite{saito2021, ma2020cape} or separate garment layers over the human body \cite{pons2017clothcap, tiwari2020sizer} with 3D scans of a clothed human obtained from body scanners. An energy-minimization framework is deployed, with objectives that combine data error and regularization terms. Although these methods successfully capture the body shapes with tight and stretchy clothes, they struggle with cases like flowing skirts that deviate significantly from the template body. \cite{Chen20TightCap} tries to remedy the problem by employing a multi-stage alignment scheme that progressively optimizes the vertex displacements. However, the consecutive intermediate optimizations introduce complexity, which may hinder reproducibility. 

Learning-based approaches can also be used for 3D garment registration. The specific cloth parameters are optimized through the pretrained neural networks to fit the template mesh to the 3D target garment mesh, by predicting the vertex-wise displacements \cite{patel20tailornet, jiang2020bcnet}. Nevertheless, the predictions are influenced by dataset biases and often lack sufficient fitting precision.

Another strategy involves fitting garments by adjusting their 2D sewing patterns, which can be optimized using differentiable simulation\cite{yu2024inverse, li2023diffavatar}. While these approaches produce simulation-ready garments, they are computationally expensive as a complete quasi-static simulation is executed at each optimization iteration. 
\section{Method}

\begin{figure}[htp]
\centering
\includegraphics[width=1.0\linewidth]{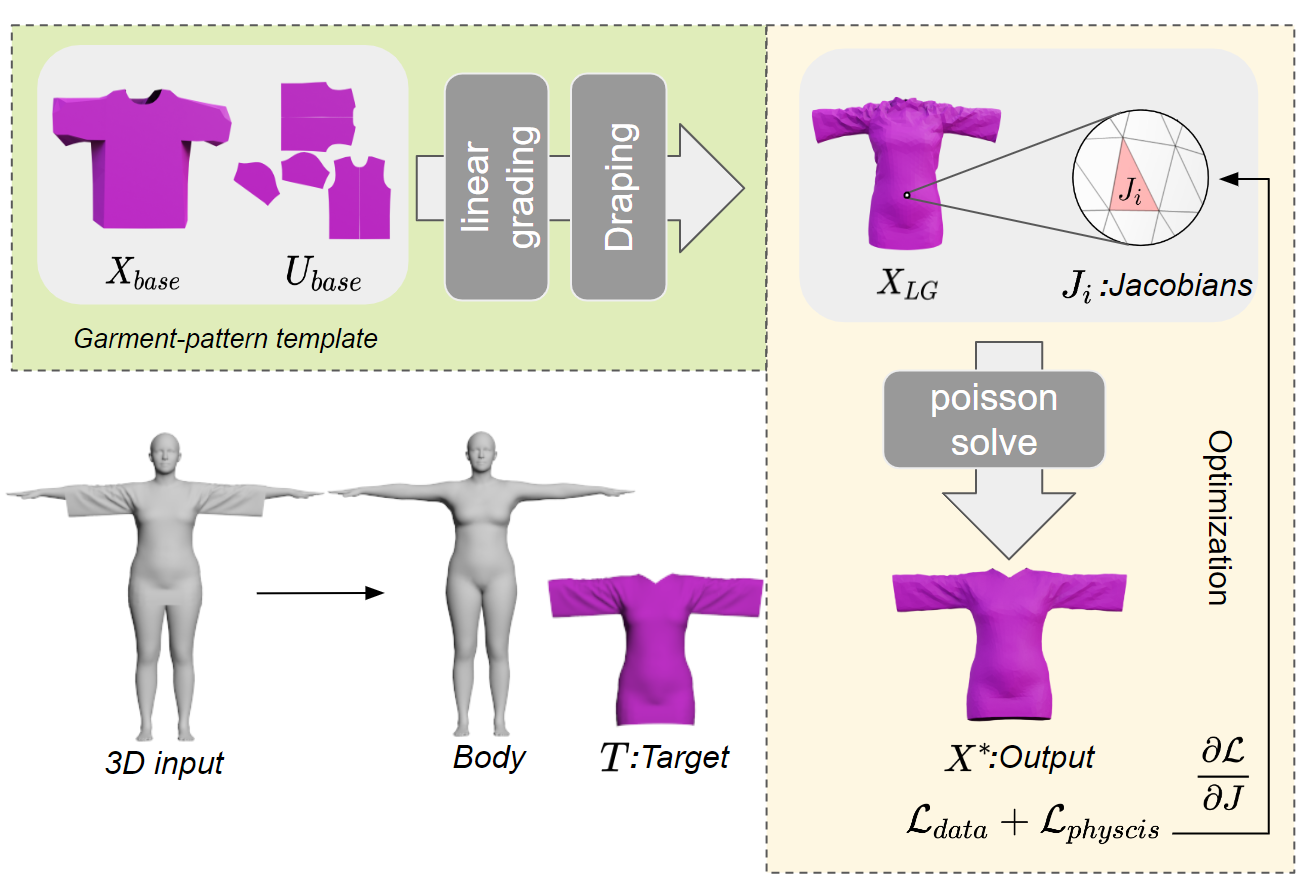}
\caption{
Given an input 3D base mesh and a target garment, we first perform linear grading on the base mesh to achieve an initial alignment. It is further refined by optimizing per-triangle Jacobians.} 
\label{fig:overview}
\end{figure}

\Cref{fig:overview} illustrates an overview of our method. PhyDeformer registers a given template mesh geometry to the target garment mesh through a two-stage process. In the first stage, the linear grading module followed by draping captures the overall geometry including size and proportions. In the second stage, a refinement is achieved by optimizing a displacement mapping $\phi: R^3 \to R^3$ over the vertices through Jacobians guided by a set of losses.

\subsection{Coarse grading}
Given a template garment mesh $X_{base}$ and its 2D sewing pattern $U_{base}$, the initial draping shape $X_{init}$ is simulated on the underlying body $B$. Note that the difference between the resulting mesh $X_{init}$ and the target $T$ may be substantial at this stage. To bridge the cap, a coarse geometric deformation is performed to capture the overall geometric shape of the target garment, such as length and proportion, by using a set of key measurements on 3D open contours. A 3D open contour is composed of edges connected to only one adjacent triangle, which often carry design features, representing elements such as necklines, hem contours, cuffs, etc. These open contours are extracted from both the draped and target meshes, each paired with its counterpart. As in \cite{yu2024inverse}, longitudinal distances between corresponding contours are measured along the body skeleton, together with the circumference differences, to adjust the 2D sewing pattern. The updated 2D pattern is then remeshed and re-draped to produce $X_{LG}$, a simulated shape that reflects the changes of 2D pattern in 3D geometry. Readers may refer to \cite{yu2024inverse} for more technical details.

\subsection{Non-rigid geometry refinement}
We further refine the simulated garment $X_{LG}$ (referred to as the source mesh hereafter) obtained from the previous stage through mesh deformation. A straightforward approach to deformation involves directly optimizing the coordinates of the source mesh vertices. However, with this method, each vertex's gradient influences only its own displacement, which can result in the excessive exposure of high-frequency details and lead to undesirable artifacts. Aiming to preserve the structure and topology of the source mesh while achieving the deformation process in a single step, we opt for deforming the vertices $X$ of the garment mesh indirectly using a set of per-triangle Jacobians (face gradients), inspired by \cite{gao2023textdeformer, aigerman2022neural}. 
In this setting, the objective is to find the optimal deformation map that captures the target shape. Starting with the variables $\mathbf{J}_i \in R^{3x3}$ initialized with per-triangle Jacobians of the source mesh, 
an optimization process updates the Jacobian field using multiple loss terms, detailed below.

As the optimization is applied to $\mathbf{J}_i$ in a per-triangle manner, a 
Poisson equation is solved to find new vertex positions well connected $\phi: R^{3} \rightarrow R^{3}$, such that the Jacobians $\nabla_{i}(\phi)$ of this deformation mapping for each triangle is closest to $\mathbf{J}_i$ in the least square sense.
\begin{equation}
\phi^{*}=\min _{\Phi} \sum_{i=1}^{|F|} A_i \left\|\nabla_{i}(\phi)-\mathbf{J}_{i}\right\|^{2},   
\end{equation}
where $A_i$ is the face area, $\nabla_{i}$ is defined as the face gradient operator, $\nabla_{i}(\phi)$ denotes the Jacobian of $\phi$ at triange $f_i$. Compared to vertex-wise optimization, the propagated gradients $\frac{\partial L}{\partial J} $ influence a larger surface mesh area, leading to a globally-coherent deformation $\phi$ which is indirectly optimized by $\mathbf{J}_i$. 

\noindent\textbf{Losses.} At each iteration, the deformed garment geometry $X \gets \phi^{*}(X)$ is compared with the target $T=\{t\}$, and the Adam algorithm is used to minimize a loss function that drives the deformation of the garment mesh. We employ a set of loss terms to ensure the refined garment mesh conforms to the target while maintaining physical realism. Several of these loss terms, specifically those arising from membrane strain energy and bending energy ($\mathcal{L}_{s}$, $\mathcal{L}_{b}$) are inspired by prior work, notably SNUG \cite{santesteban2022snug}. 

We use a reconstruction loss to evaluate the similarity between $X$ and $T$. This loss comprises Chamfer distances $\mathcal{L}_{CF}$ calculated between both the surfaces and the open contours,
\begin{equation}
\mathcal{L}_{rec}=\mathcal{L}_{CF}(X, T)+\mathcal{L}_{CF}(X_{open},T_{open}).  
\end{equation}

The cosine distance of normals $\mathcal{L}_n$ between $X$ and $T$ is also measured, which is written as:
\begin{equation}
\mathcal{L}_{n}=\frac{1}{|X|} \sum_{x}^{|X|}\left(1-\left\langle\mathbf{n}_{x}, \mathbf{n}_{\tilde{t}}\right\rangle\right)+\frac{1}{|T|} \sum_{t}^{|T|}\left(1-\left\langle\mathbf{n}_{t}, \mathbf{n}_{\tilde{x}}\right\rangle\right),
\end{equation}
where $\mathbf{n}_{x}$ and $\mathbf{n}_{\tilde{t}}$ are the unit normal vectors at point $x$ and $\tilde{t}=\arg \min_{t \in T} {(\|x-t\|)} $ respectively, and vice versa for $\mathbf{n}_{t}$ and $\mathbf{n}_{\tilde{x}}$.

Strain loss is employed to ensure that the shapes of the triangles in the deformed garment resist stretching. This loss term is based on the Saint Venant Kirchhoff (StVK) elastic material model, which is formulated as:
\begin{equation}
\mathcal{L}_{s}=\sum^{|F|}_i \left(
\frac{\lambda}{2}\cdotp 
tr\left(\mathbf{G}_i\right)^{2}+
\mu \cdotp tr\left(\mathbf{G}_i^{2}\right)
\right)A_i,
\end{equation}
where $\lambda$ and $\mu$ represent the Lamé coefficients, which are respectively set to 16.3 and to 13.5 in our experiments. $A_i$ is the area of $i$-th triangle, and $\mathbf{G_i}$ the green strain tensor. This regularization term constrains the deformation, ensuring that the shapes of the triangles do not deviate excessively from the original, undeformed geometry.

Bending loss is employed to penalize changes in discrete curvature as a function of the dihedral angle between edge-adjacent triangles, which is formulated as follows:
\begin{equation}
\mathcal{L}_{b}=\sum^{|E|}_j \frac{\kappa}{2} \cdotp \alpha^{2}_j, 
\end{equation}
where $\kappa$ represents the bending stiffness set to $4e-5$, $E$ is the edges, and $\alpha_j$ the radian angle between two adjacent triangles. This loss term effectively constrains the deformation of the garment to prevent excessive bending.

Finally, collision loss is employed to prevent the interpenetration of garment vertices with the body mesh (if available). This is achieved by penalizing the negative distance between garment nodes and their closest point on the body surface with a cubic energy term:
\begin{equation}
\mathcal{L}_{c}=\sum^{|X|}_n \max (\epsilon-sdf(x), 0)^{3},
\end{equation}
where $sdf(\cdotp)$ represents the distance between the query point and the body surface, and $\epsilon$ the chosen minimal distance threshold between the body and the garment. This loss term constrains the deformation of the garment to prevent the intersection with the body mesh, leading to a more physically correct outcome.

All of these loss terms are combined to form the final objective
to guide the deformation of the source mesh
:
\begin{equation}
\mathcal{L}= \mathcal{L}_{rec}+\lambda_{n} \mathcal{L}_{n}+\lambda_{s} \mathcal{L}_{s}+\lambda_{b} \mathcal{L}_{b}+\lambda_{c} \mathcal{L}_{c}.
\end{equation}

\section{Results and Experiments}

\subsection{Implementation details} 
We conducted the experiments using a NVIDIA 3090 GPU, 24Gb RAM, and an Intel i7-5220R CPU. We run the optimization for 1500 iterations until the convergence is reached, which takes approximately 5 minutes. The stretching regularization term is added after 500 iterations. The learning rate was set to 0.002 in the Adam optimizer. The loss weights were set as: 
 $\lambda_{n}=0.01$, 
 $\lambda_{s}=1$,
 $\lambda_{b}=0.1$, and
 $\lambda_{c}=0.01$ in our experiments. We use template base models selected from the Berkeley Garment Library\cite{narain2012adaptive}.
\subsection{Evaluation on synthetic garments}
\begin{figure}[htp]
\centering
\includegraphics[width=1.0\linewidth]{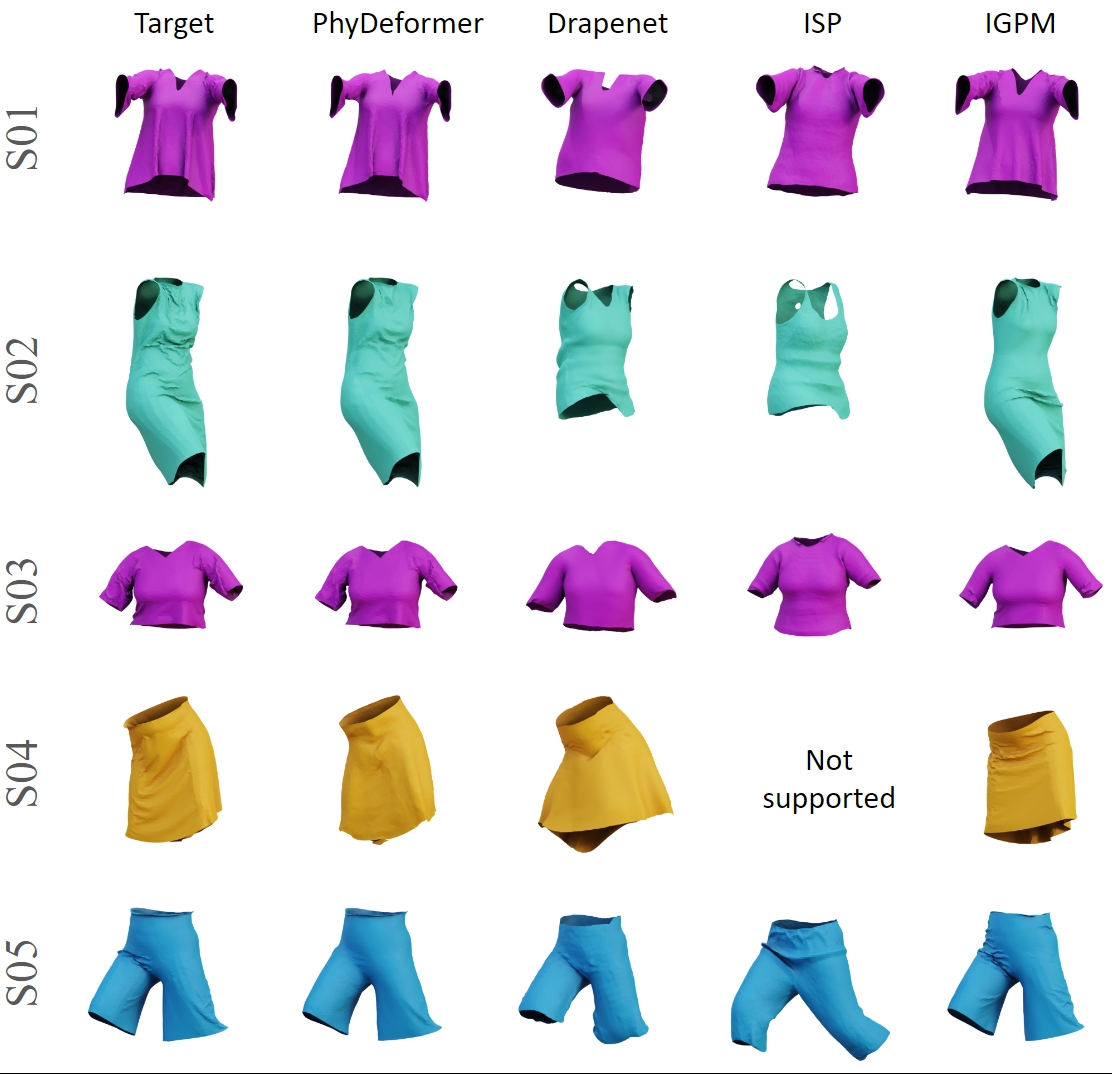}
\caption{
Qualitative comparison of 3D garment reconstruction using our method versus other approaches \cite{yu2024inverse, de2023drapenet, li2024isp}. 
} 
\label{fig:3deval2}
\end{figure}

To demonstrate the effectiveness of our garment registration approach, we tested on samples from the Sewfactory dataset\cite{liu23sewformer} containing diverse garment styles and shapes in different poses.  We then qualitatively evaluated the results with state-of-the-art works as shown in \Cref{fig:3deval2}. Drapenet\cite{de2023drapenet} uses unsigned distance functions (UDFs) and requires extra computations for meshing, and it is sensitive to the initialization of latent code for optimization. Both ISP\cite{li2024isp} and Drapenet can produce certain geometric details in the registration to garments, but when it comes to challenging posed garment samples such as S01 and S05, the performance decreases and leads to visible defects. IGPM\cite{yu2024inverse} exploits the inverse cloth simulation to achieve coarse-to-fine alignment, but it fails to capture the exact wrinkle patterns and the inverse simulation is relatively expensive to execute. In contrast, PhyDeformer reconstructs accurate 3D geometry, for both loose and tight garments of different subjects with less computational cost. For quantitative evaluation of the geometric similarity, two metrics are used: Chamfer distance(L2 norm, scaled by $e^3$) to the ground truth mesh vertices, and the normal similarity to measure the orientation consistency of the surface. As shown in \Cref{tab:3drecon2}, our method outperforms others in the 3D reconstruction of posed garments.

\begin{table}[ht!]
\centering
\caption{Quantitative evaluation in 3D garment reconstruction. 
}
\renewcommand{\arraystretch}{1.1}
\scalebox{0.8}{
\begin{tabular}{ccccc}
\noalign{\smallskip}
      & \multicolumn{4}{c}{Results on Sewfactory dataset}\\
\noalign{\smallskip}
      & \multicolumn{4}{c}{Chamfer distance / Normal similarity} \\
Garments & Drapenet  & ISP & IGPM  & Ours \\
\hline
S01  & 0.346 / 0.235   & 0.431 /0.232 & 0.191 / 0.203 &\textbf{0.159} / \textbf{0.058}
\\
S02 & 17.031 / 0.165   & 17.61 /0.214 & 0.083 / 0.089 & \textbf{0.047} / \textbf{0.021}
\\
S03 &  0.798 / 0.156   & 0.182 / 0.164 & 0.205 / 0.124 & \textbf{0.076} / \textbf{0.061}
\\
S04 &    -    /  -     & 3.024 / 0.340 & 0.398 / 0.276 &\textbf{0.228} / \textbf{0.27}
\\
S05 &  1.831/ 0.268    & 0.87 / 0.186 & 0.090 / 0.07 &\textbf{0.138} / \textbf{0.028}
\\
\noalign{\smallskip}
\hline
\end{tabular}
}
\label{tab:3drecon2}
\end{table}



\subsection{Evaluation on 3D scan data}
\begin{figure}[htp]
\centering
\includegraphics[width=1.0\linewidth]{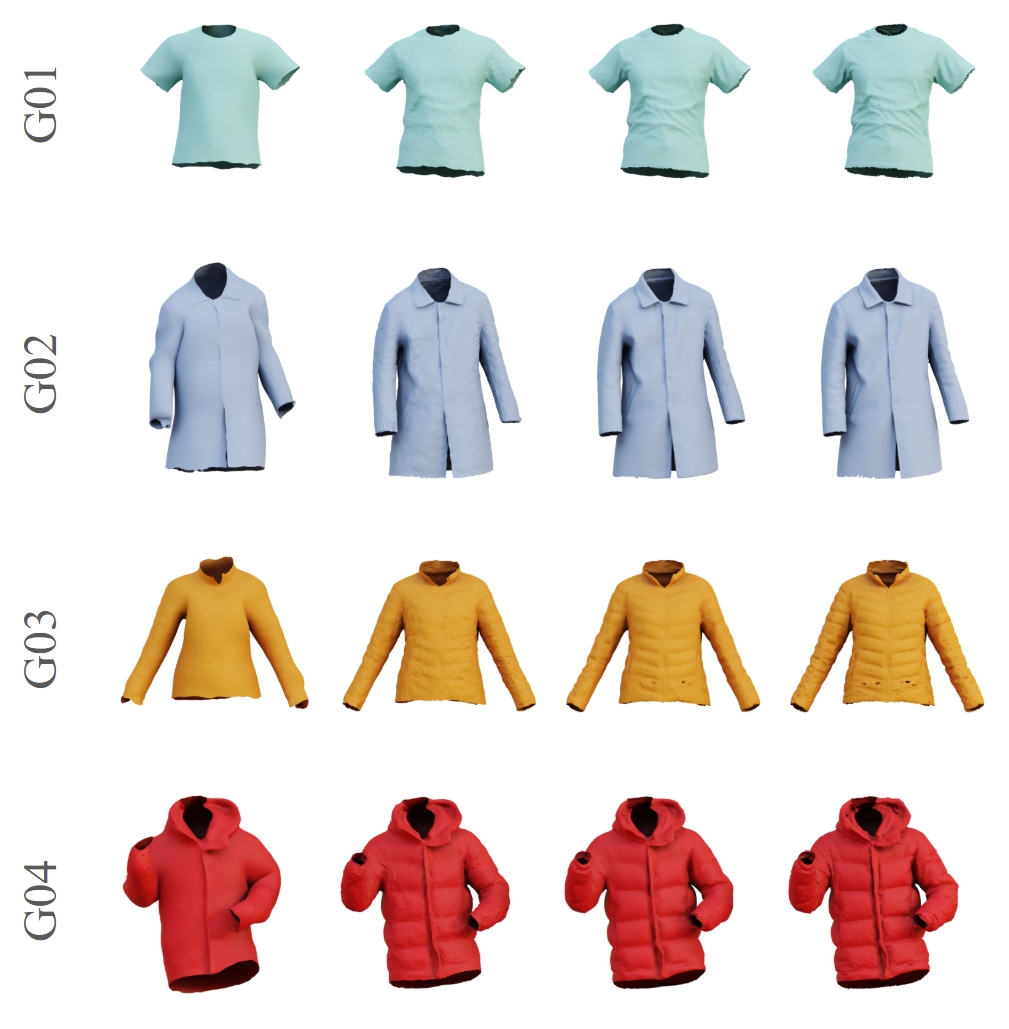}
\caption{
Qualitative results on GarmCap dataset. Leftmost column: posed templates, second column: coarse fittings, third column: refined fittings by PhyDeformer, Rightmost column: target shape. Best viewed zoomed-in.}
\label{ch2:fig:scan}
\end{figure}

To showcase the capability of our refinement stage in handling fine-grained, wrinkle-level details from real scans, as well as its ease of integration with other coarse fitting techniques, we performed a qualitative evaluation using the GarmCap dataset \cite{lin2023}. 
We adopted their rigged smooth template and coarse fitting, refining the alignment using our second step. As shown in \Cref{ch2:fig:scan}, our method outputs reasonable, high-quality reconstruction of the 3D scanned garments.

\subsection{Use case: Registration for inverse garment simulation}
\begin{figure}[htp]
\centering
\includegraphics[width=1.0\linewidth]{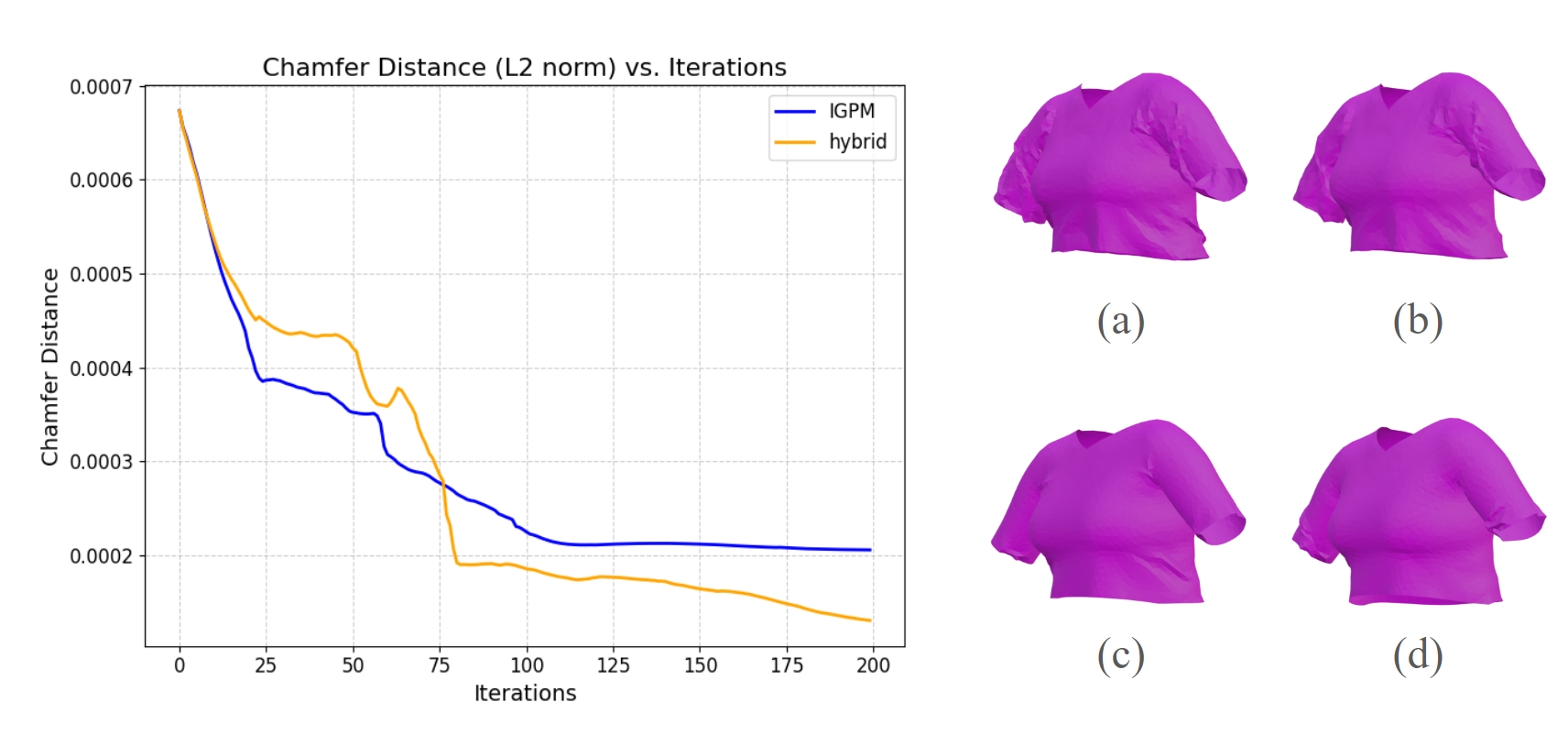}
\caption{We evaluate the Chamfer Distance error for hybrid and IGPM \cite{yu2024inverse}. On the right, we also illustrate the qualitative results for comparison: (a) Target; (b) New target by PhyDeformer; (c) IGBM with Hybrid scheme, and (d) IGPM.}
\label{fig:hybrid}
\end{figure}

PhyDeformer offers significantly better efficiency than the fully physics-based simulation-embedded method IGPM\cite{yu2024inverse}, whereas the latter offers simulation-ready assets for more flexible reusability. PhyDeformer can provide a registered target mesh with consistent topology. After the linear grading stage, the output is registered to the raw target mesh, the resulting deformed mesh is utilized as a new target for the differentiable simulation optimization, replacing the Chamfer distance loss with the per-vertex loss (MSE, mean square error). We refer to this scheme as \textit{Hybrid} in contrast to the baseline IGPM. In \Cref{fig:hybrid}, we showcase how the hybrid strategy outperforms the vanilla IGPM in terms of the speed of convergence. Since MSE is more expressive, it also leads to more accurate reconstruction.



\section{Ablation study}
We conduct ablation studies to evaluate how the components affect the overall fitting quality. 
Some results are shown in \Cref{tab:ablation} for the garment S01, the most representative data in our experiments. 

\begin{figure}[htp]
\centering
\includegraphics[width=1.0\linewidth]{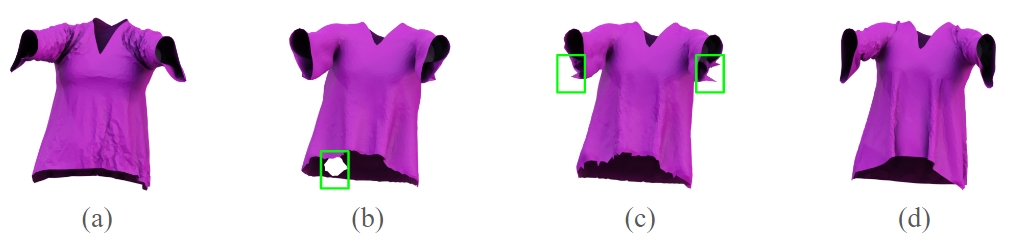}
\caption{
Results of optimizing vertex displacements instead of Jacobians. From left to right: (a) source mesh, (b) deformed mesh with naive vertex-displacements optimization, (c) deformed mesh with clipped gradients to avoid holes caused by "NaNs", (d) target mesh.}
\label{fig:ablation_Jacobians}
\end{figure}

\noindent\textbf{Without Jacobians.} 
In \Cref{fig:ablation_Jacobians}, we show
that replacing Jacobians with vertex displacements in the optimization decreases the resulting surface quality. Representing the deformation through vertex displacements exposes the per-vertex high-frequency mode of the deformation, thus leading to localized noisy gradients that deteriorate the triangulation of the meshes(holes) as illustrated in \Cref{fig:ablation_Jacobians} (b). Clipping the gradients can alleviate this but still causes sharp "burrs" on the deformed mesh shown in \Cref{fig:ablation_Jacobians} (c).

\begin{figure}[htp]
\centering
\includegraphics[width=0.8\linewidth]{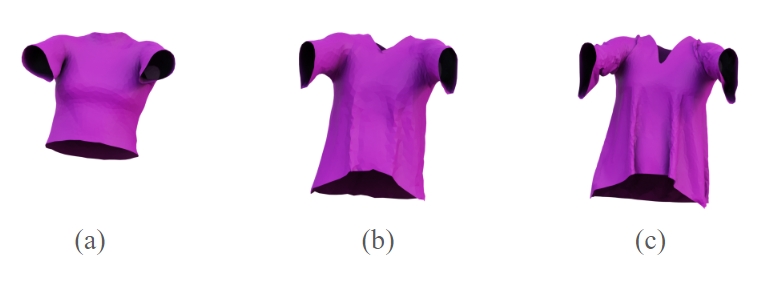}
\caption{Results obtained without linear grading: (a) posed mesh without linear grading, (b) deformed mesh, (c) target mesh.}
\label{fig:ablation_lg}
\end{figure}

\noindent\textbf{Without linear grading.} In \Cref{fig:ablation_lg} (b) we illustrate that our method's performance decreases when the linear grading stage is disabled. 
The linear grading provides better initialization, thereby contributing to improved overall performance.

\begin{figure}[htp]
\centering
\includegraphics[width=1.0\linewidth]{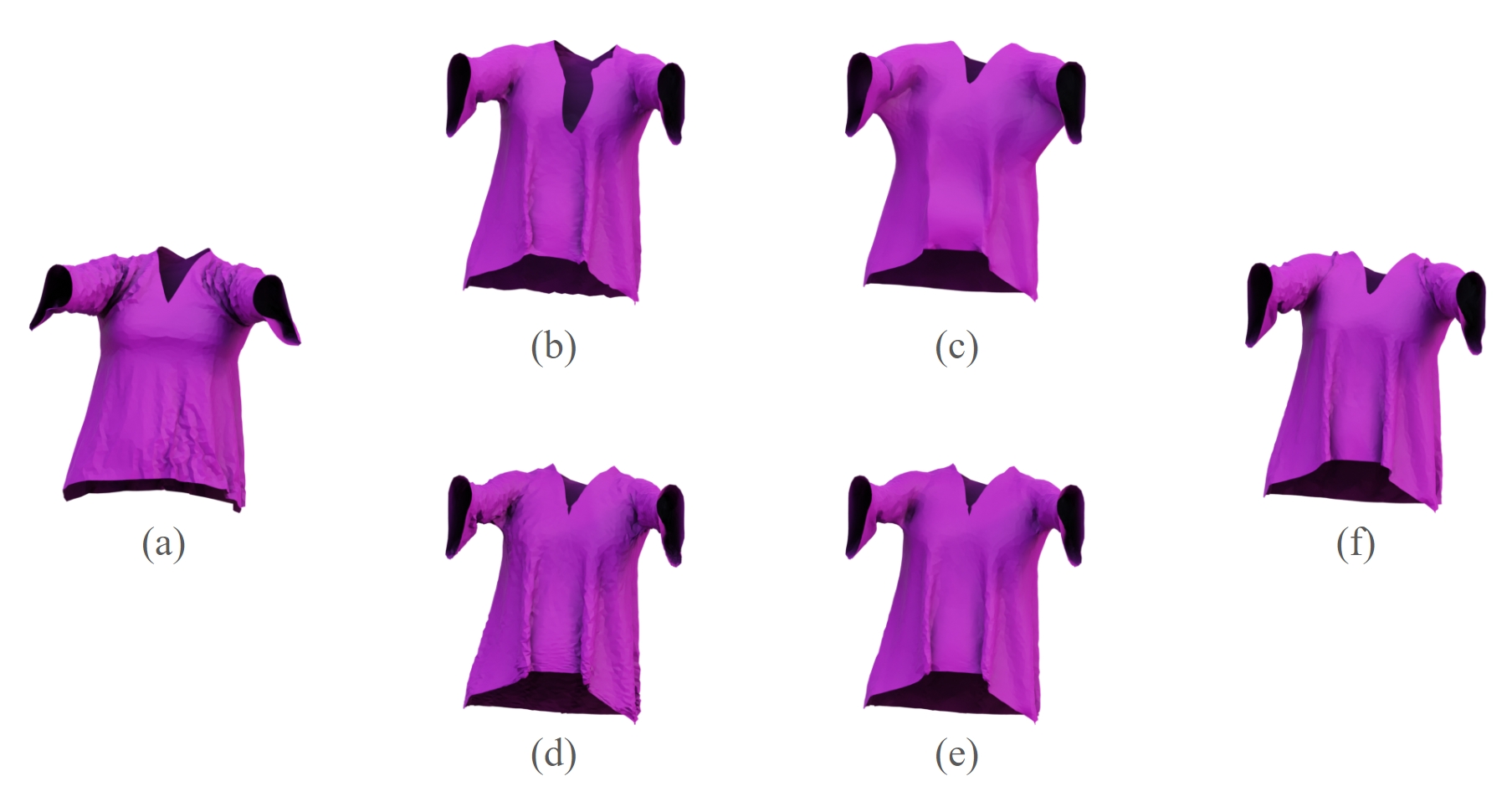}
\caption{Results of our ablation study on loss terms: (a) source mesh, (b) without contour loss, 
 (c) without normal loss, (d) without bending loss, (e)with all losses, (f) target mesh.}
\label{fig:ablation3}
\end{figure}

\noindent\textbf{Without loss terms.}
\Cref{fig:ablation3} shows that our refinement stage with all losses can lead to large and smooth deformations for the final fitting. To assess the effectiveness of each loss term, we conducted an ablation study by omitting individual losses — specifically, the contour loss, normal loss, and bending loss.


As shown in \Cref{tab:ablation}, the removal of the open contour loss term did not affect much the global geometric shape, though it led to an increase in Chamfer distance. The consequence is more perceptible in \Cref{fig:ablation3} (b), where a visible failure around the collar region can be observed. Eliminating the normal restriction significantly compromised the resulting deformed mesh, as demonstrated in \Cref{fig:ablation3} (c). Similarly, omitting the bending loss introduced evident bumpy artifacts, shown in \Cref{fig:ablation3} (d). Quantitative results in \Cref{tab:ablation} confirm that our two-phase refining strategy enhances the geometric accuracy of the output.

\begin{table}[htb!]
\caption{Quantitative results of ablation studies. We report the
metrics for the garment S01. The best results are in boldface.
}
\scalebox{0.9}{
\centering
\begin{tabular}{ lcc }
\hline
Method & Chamfer distance (CF)    & Normal similarity\\
\hline
w/o linear grading & 0.188 &  0.106\\
w/o contour loss & 0.088 & 0.056 \\
w/o normal consistency & 0.124 & 0.173  \\
w/o bending loss & 0.158 & 0.052\\
Ours& \textbf{0.047} & \textbf{0.021}\\
\hline 
\end{tabular}
}
\label{tab:ablation}
\end{table}

\section{Robustness to noisy targets}
\begin{figure}[htp]
\centering
\includegraphics[width=1.0\linewidth]{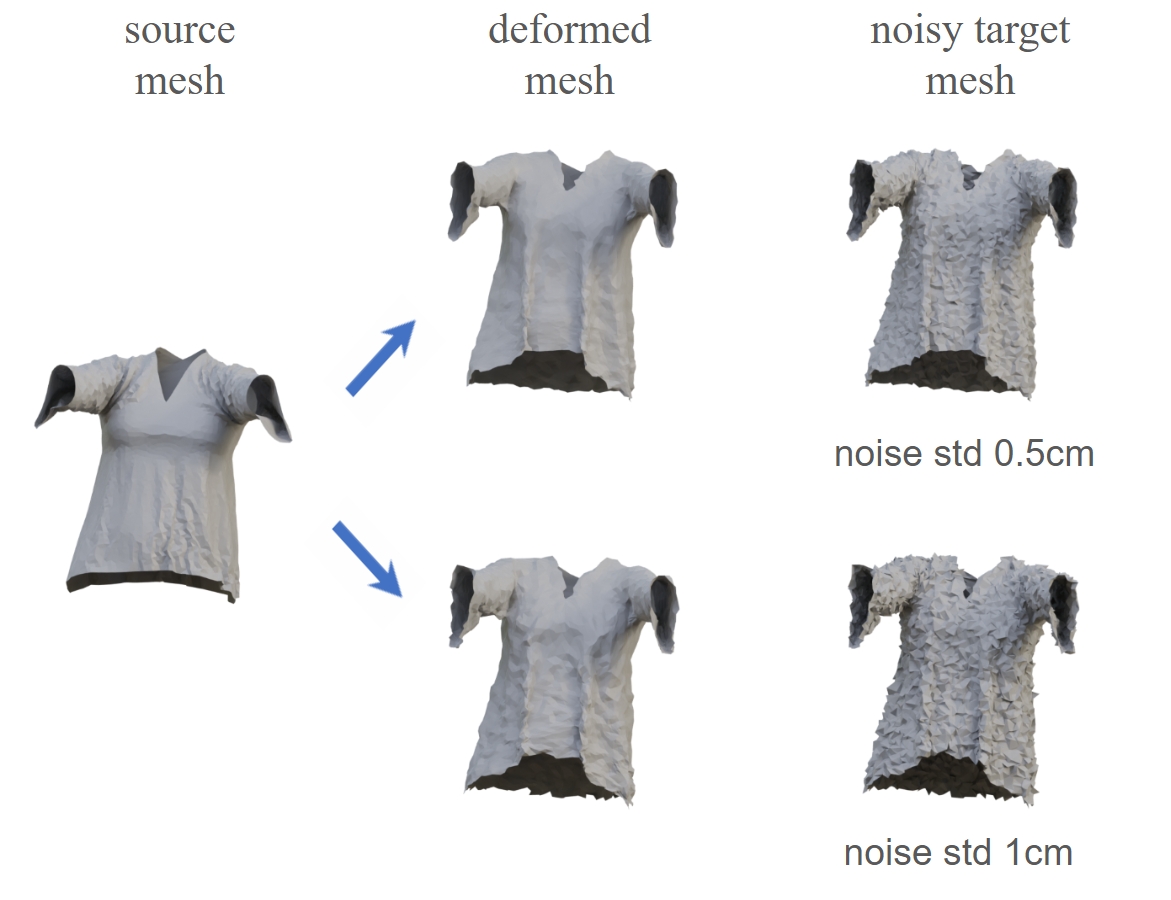}
\caption{The alignment results on target meshes with synthetic noise of two different levels. 
}
\label{fig:noise}
\end{figure}

We assess its robustness against noise present in the target shape. We simulate the noisy artifacts by introducing Gaussian noise to the vertices coordinates of the target mesh, adjusting the standard deviation between 0.5 and 1 cm. When evaluating PhyDeformer with these modified targets (see \Cref{fig:noise}), we observe a decline in performance as noise levels increase. This indicates that PhyDeformer can effectively manage minimal noise levels, typical of high-accuracy 3D scanning systems.

\section{Conclusion and limitation}
In this work, we introduce PhyDeformer, which is lightweight yet capable of producing high-quality registered meshes with detailed wrinkles. Physics-driven loss terms are employed to constrain Jacobian-based deformation, ensuring approximate adherence to physically faithful garment shapes.

\bibliographystyle{eg-alpha-doi} 
\bibliography{egbibsample}       

\newcommand{\etalchar}[1]{$^{#1}$}
\begin{thebibliography}{\uppercase{TBTPM20}}

\bibitem[AGK{\etalchar{*}}22]{aigerman2022neural}
\textsc{Aigerman N., Gupta K., Kim V.~G., Chaudhuri S., Saito J., Groueix T.}:
\newblock Neural jacobian fields: Learning intrinsic mappings of arbitrary meshes.
\newblock \emph{SIGGRAPH} (2022).

\bibitem[CPY{\etalchar{*}}21]{Chen20TightCap}
\textsc{Chen X., Pang A., Yang W., Wang P., Xu L., Yu J.}:
\newblock Tightcap: 3d human shape capture with clothing tightness field.
\newblock \emph{ACM Trans. Graph. 41}, 1 (nov 2021).
\newblock \href {https://doi.org/10.1145/3478518} {\path{doi:10.1145/3478518}}.

\bibitem[DLLG{\etalchar{*}}23]{de2023drapenet}
\textsc{De~Luigi L., Li R., Guillard B., Salzmann M., Fua P.}:
\newblock Drapenet: Garment generation and self-supervised draping.
\newblock In \emph{Proceedings of the IEEE/CVF Conference on Computer Vision and Pattern Recognition} (2023), pp.~1451--1460.

\bibitem[GAG{\etalchar{*}}23]{gao2023textdeformer}
\textsc{Gao W., Aigerman N., Groueix T., Kim V., Hanocka R.}:
\newblock Textdeformer: Geometry manipulation using text guidance.
\newblock In \emph{ACM SIGGRAPH 2023 Conference Proceedings} (2023), pp.~1--11.

\bibitem[JZH{\etalchar{*}}20]{jiang2020bcnet}
\textsc{Jiang B., Zhang J., Hong Y., Luo J., Liu L., Bao H.}:
\newblock Bcnet: Learning body and cloth shape from a single image.
\newblock In \emph{Computer Vision--ECCV 2020: 16th European Conference, Glasgow, UK, August 23--28, 2020, Proceedings, Part XX 16} (2020), Springer, pp.~18--35.

\bibitem[LCL{\etalchar{*}}24]{li2023diffavatar}
\textsc{Li Y., Chen H.-y., Larionov E., Sarafianos N., Matusik W., Stuyck T.}:
\newblock {DiffAvatar}: Simulation-ready garment optimization with differentiable simulation.
\newblock In \emph{Proceedings of the IEEE/CVF Conference on Computer Vision and Pattern Recognition (CVPR)} (June 2024).
\newblock \href {http://arxiv.org/abs/2311.12194} {\path{arXiv:2311.12194}}.

\bibitem[LGF24]{li2024isp}
\textsc{Li R., Guillard B., Fua P.}:
\newblock Isp: Multi-layered garment draping with implicit sewing patterns.
\newblock \emph{Advances in Neural Information Processing Systems 36} (2024).

\bibitem[LXL{\etalchar{*}}23]{liu23sewformer}
\textsc{Liu L., Xu X., Lin Z., Liang J., Yan S.}:
\newblock Towards garment sewing pattern reconstruction from a single image.
\newblock \emph{ACM Transactions on Graphics (TOG) 42}, 6 (2023), 1--15.

\bibitem[LZZ{\etalchar{*}}23]{lin2023}
\textsc{Lin S., Zhou B., Zheng Z., Zhang H., Liu Y.}:
\newblock Leveraging intrinsic properties for non-rigid garment alignment.
\newblock In \emph{IEEE/CVF International Conference on Computer Vision (ICCV)} (2023).

\bibitem[MYR{\etalchar{*}}20]{ma2020cape}
\textsc{Ma Q., Yang J., Ranjan A., Pujades S., Pons-Moll G., Tang S., Black M.~J.}:
\newblock Learning to dress 3d people in generative clothing.
\newblock In \emph{Proceedings of the IEEE/CVF Conference on Computer Vision and Pattern Recognition} (2020), pp.~6469--6478.

\bibitem[NSO12]{narain2012adaptive}
\textsc{Narain R., Samii A., O'brien J.~F.}:
\newblock Adaptive anisotropic remeshing for cloth simulation.
\newblock \emph{ACM transactions on graphics (TOG) 31}, 6 (2012), 1--10.

\bibitem[PLPM20]{patel20tailornet}
\textsc{Patel C., Liao Z., Pons-Moll G.}:
\newblock Tailornet: Predicting clothing in 3d as a function of human pose, shape and garment style.
\newblock In \emph{{IEEE} Conference on Computer Vision and Pattern Recognition (CVPR)} (jun 2020), {IEEE}.

\bibitem[PMPHB17]{pons2017clothcap}
\textsc{Pons-Moll G., Pujades S., Hu S., Black M.~J.}:
\newblock Clothcap: Seamless 4d clothing capture and retargeting.
\newblock \emph{ACM Transactions on Graphics (ToG) 36}, 4 (2017), 1--15.

\bibitem[SOC22]{santesteban2022snug}
\textsc{Santesteban I., Otaduy M.~A., Casas D.}:
\newblock Snug: Self-supervised neural dynamic garments.
\newblock In \emph{Proceedings of the IEEE/CVF Conference on Computer Vision and Pattern Recognition} (2022), pp.~8140--8150.

\bibitem[SYMB21]{saito2021}
\textsc{Saito S., Yang J., Ma Q., Black M.~J.}:
\newblock {SCANimate}: Weakly supervised learning of skinned clothed avatar networks.
\newblock In \emph{Proceedings IEEE/CVF Conf.~on Computer Vision and Pattern Recognition (CVPR)} (June 2021).

\bibitem[TBTPM20]{tiwari2020sizer}
\textsc{Tiwari G., Bhatnagar B.~L., Tung T., Pons-Moll G.}:
\newblock Sizer: A dataset and model for parsing 3d clothing and learning size sensitive 3d clothing.
\newblock In \emph{Computer Vision--ECCV 2020: 16th European Conference, Glasgow, UK, August 23--28, 2020, Proceedings, Part III 16} (2020), Springer, pp.~1--18.

\bibitem[YCS24]{yu2024inverse}
\textsc{Yu B., Cordier F., Seo H.}:
\newblock Inverse garment and pattern modeling with a differentiable simulator.
\newblock In \emph{Computer Graphics Forum} (2024), Wiley Online Library, p.~e15249.

\end{thebibliography}


\end{document}